\documentclass[11pt,a4paper]{article}
\usepackage{acl2021}
\usepackage{times}
\usepackage{latexsym}

\usepackage{times}
\usepackage{adjustbox}
\usepackage{latexsym}
\usepackage{booktabs}
\usepackage{tabularx}
\usepackage{multirow}
\usepackage{amsmath}
\usepackage{xcolor}
\usepackage{algorithm}
\usepackage{algorithmic}
\usepackage{enumitem}
\usepackage{caption}
\usepackage{subcaption}

\usepackage[T1]{fontenc}

\usepackage[utf8]{inputenc}
\usepackage{microtype}

\title{Denoising Enhanced Distantly Supervised Ultrafine Entity Typing}

\author{\vspace{0.1in}\\\textbf{Yue Zhang, Hongliang Fei, Ping Li} \\\\
Cognitive Computing Lab\\
Baidu Research\\
10900 NE 8th St. Bellevue, WA 98004, USA\\
  \texttt{\{yuezhang030, feihongliang0,  pingli98\}@gmail.com}
}

\begin{document}
\maketitle
\begin{abstract}
Recently, the task of distantly supervised (DS) ultra-fine entity typing has received significant attention. However, DS data is noisy and often suffers from missing or wrong labeling issues resulting in low precision and low recall. This paper proposes a novel ultra-fine entity typing model with denoising capability. Specifically, we build a noise model to estimate the unknown labeling noise distribution over input contexts and noisy type labels. With the noise model, more trustworthy labels can be recovered by subtracting the estimated noise from the input. Furthermore, we propose an entity typing model, which adopts a bi-encoder architecture, is trained on the denoised data. Finally, the noise model and entity typing model are trained iteratively to enhance each other. We conduct extensive experiments on the Ultra-Fine entity typing dataset as well as OntoNotes dataset and demonstrate that our approach significantly outperforms other baseline methods.
\end{abstract}

\section{Introduction}

Entity typing is the task of identifying specific semantic types of entity mentions in given contexts. Recently, more and more research has focused on ultra-fine entity typing~\cite{choi2018ultra, onoe2019learning, dai2021ultra}. Comparing to traditional entity typing tasks~\cite{ren2016afet, ren2016label, xu2018neural, ling2012fine, yosef2013hyena, abhishek2017fine, shimaoka2017neural, xin2018put}, the type set in ultra-fine entity typing is not restricted by KB schema, but includes a vast number of free-form types. 

To automatically annotate the large-scale ultra-fine entity typing data,~\citet{choi2018ultra} utilized different sources for distant supervision (DS), including: 1) entity linking, where they mine entity mentions that were linked to Wikipedia in HTML,
and extract relevant types from their encyclopedic definitions, and 2) head words, where they automatically extracted nominal head words from raw text as types. However, distant supervision often suffers from the low-precision and low-recall problems~\cite{ren2016label}, where recall can suffer from KB or Wikipedia incompleteness, and precision can suffer when the selected types do not fit the context.

\begin{table}[ht]
\vspace{0.1in}
    \centering
    \begin{tabular}{p{5cm}|p{1.75cm}}
    \toprule
    Instance & DS label \\
    \midrule
    \texttt{S1}: On her first match on grass at the AEGON International in Eastbourne, Lisicki lost to [\underline{Samantha Stosur}] in the first round. & \textcolor{red}{actor}, athlete, person \\
    \midrule
    \texttt{S2}: [\underline{The film}] was adapted by Hugh Walpole, Howard Estabrook and Lenore J. Coffee from the Dickens novel, and directed by George Cukor. & film, \textcolor{gray}{movie, show, art, entertainment, creation} \\
    \bottomrule
    \end{tabular}
    \caption{Examples selected from the Ultra-Fine Entity Typing dataset in ~\citet{choi2018ultra}. }
    \label{table:example}
\end{table}

Table~\ref{table:example} shows two examples from these datasets~\cite{choi2018ultra} to illustrate the challenges in automatic annotation using distant supervision. Sentence \texttt{S1} is incorrectly annotated as \texttt{actor} through entity linking, which is beyond the given context. Sentence \texttt{S2} shows that simply treating the head word \texttt{film} as the type label, while correct in this case, but misses many other valid types: \texttt{movie}, \texttt{show}, \texttt{art}, etc.

To address the noisy labeling problem in distantly supervised entity typing, researchers devoted much effort to denoising.~\citet{xiong2019imposing} learns the hierarchical correlations between different types by injecting type co-occurrence Graph.~\citet{onoe2021modeling} considers box embedding, which is more robust to data noise. While these methods implicitly learn to denoise data noise, it is difficult for humans to interpret their denoising capacity. ~\citet{onoe2019learning} proposed an explicit denoising method, where they learn a filtering function and a relabeling function to denoise DS data and then train an entity typing model on the denoised DS dataset. However, they only utilized a small scale gold data to learn the filtering and relabeling function. Besides, their model did not model the dependency between context and entity phrases.

In this paper, we aim to develop an explicit denoising method for distantly supervised ultra-fine entity typing. Our framework mainly consists of two modules: a noise modeling component and an entity typing model. The noise model estimates the unknown labeling noise distribution over input contexts and observed (noisy) type labels. However, noise modeling is challenging because the noise information in the DS data is often unavailable, and noise can vary with different distant labeling techniques. To model the noise, we perturb the small-scale gold-labeled dataset's labels to mimic the DS's noise. Additionally, we utilize the $L_1$ norm regularization on the large-scale DS data to pursue the sparseness of labeling noise. Our noise model conditions on the input context sentence and its noisy labels to measure the underlying noise, where the denoised data can be recovered from DS data by subtracting the noise. For the entity typing model, we adopt a bi-encoder architecture to match input context and type phrases and train the entity typing model on gold labeled and denoised data. Finally, we design an iterative training~\cite{tanaka2018joint, xie2020self} procedure to train the noise model and entity typing model iteratively to enhance each other. 

\vspace{0.1in}

We summarize our \textbf{contributions} as follows:

\vspace{0.05in}
\noindent(i) We propose a denoising enhanced ultra-fine entity typing model under the distant supervised setting, including noise modeling and entity typing modeling. Unlike previous denoising work~\cite{onoe2019learning} to filter low-quality samples, our noise model directly measures underlying labeling noise, regardless of DS techniques.

\vspace{0.05in}
\noindent(ii) \citet{onoe2019learning} learns a relabel function to directly relabel samples, while, we model the labeling noise. iii) We evaluate our model on both the Ultra-Fine entity typing (UFET) and OntoNotes datasets, which are benchmarks for distantly supervised ultra-fine entity typing and fine-grained entity typing tasks. We show that our model can effectively denoise the DS data and learn a superior entity typing model through detailed comparison, analysis, and case study.

\section{Related Works}

\begin{figure*}[!htb]
\centering
\includegraphics[width=1\textwidth]{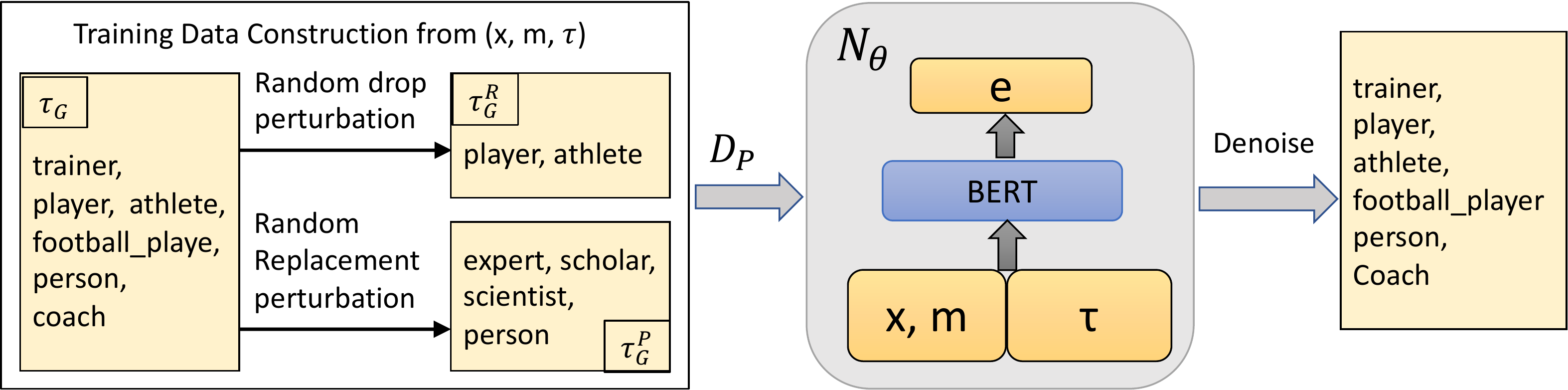} 
\caption{The procedure of training our noise model using one example. We use an instance from gold dataset ``But Laporte, who names his World Cup squad on Wednesday, feels [\underline{he}] has now found a World Cup fly...'' as example, where we perturb the gold type set $\tau_G$ into low-recall set $\tau_G^R$ and low-precision set $\tau_G^P$, separately. The noise model $\mathcal{N}_\theta$ takes the perturbed data as input, and outputs the estimated noise. } 
\label{fig:architecture}
\end{figure*}

\subsection{Ultra-Fine Entity Typing}

The ultra-fine entity typing task was first proposed by~\citet{choi2018ultra}. They considered a multitask objective, where they divide labels into three bins (general, fine, and ultra-fine), and update labels only in a bin containing at least one positive label. To further reduce the distant supervision noise,~\citet{xiong2019imposing} introduces a graph propagation layer to impose a label-relational bias on entity typing models to implicitly capture type dependencies. ~\citet{onoe2021modeling} uses box embedding to capture latent type hierarchies, which is more robust to the labeling noise comparing to vector embedding.~\citet{dai2021ultra} proposes to obtain more weakly supervised training data by prompting weak labels from language models. ~\citet{zhang2022end} leverages retrieval augmentation to resolve the distant supervision noise. 

Among the previous works,~\citet{onoe2019learning} is the most similar to ours, where the filtering function is used to discard useless instances, and relabeling function is used to relabel an instance. Through filtering and relabeling,~\citet{onoe2019learning} explicitly denoise the distant supervision data. However, their denoising procedure is trained only on a small-scale gold-labeled data, while ignoring the large-scale data with distant supervision labels. In addition, our denoising method directly models the underlying label noise instead of brutally filtering all the samples with partial wrong labels. 

\subsection{Learning from Noisy Labeled Datasets}

We briefly review the broad techniques for learning from noisy labeled datasets. Traditionally, regularization is an efficient method to deal with the issue of DNNs easily fitting noisy labels, including weight decay, and dropout.  Besides, a few studies achieve noise-robust classification using noise-tolerant loss functions, such as mean square error and mean absolute error~\cite{ghosh2017robust}. Recently, self-training~\cite{xie2020self} first uses labeled data to train a good teacher model, then uses the teacher model to label unlabeled data, and finally uses the labeled data and unlabeled data to jointly train a student model. Furthermore, various noise modeling methods are developed, including normalizing flows based methods~\cite{abdelhamed2019noise}, and GAN based methods~\cite{chen2018image}. However, these noise modeling methods cannot be directly adapted to NLP tasks because of the differentiability issues.

\section{Methodology}

\subsection{Problem Setup}
Given $l$ gold labeled triplets (context, mention, label) $\mathcal{D}_{\textrm{G}} =\{(x_{\textrm{G}}^{(i)}, m_{\textrm{G}}^{(i)},  \mathbf{Y}_{\textrm{G}}^{(i)})\}_{i=1}^{l}$ and $n$ noisily labeled triplets $\mathcal{D}_{\textrm{N}} =\{(x_{\textrm{N}}^{(i)}, m_{\textrm{N}}^{(i)}, \mathbf{Y}_{\textrm{N}}^{(i)})\}_{i=1}^{n}$, where both $\mathbf{Y}_\textrm{G}^{(i)}$ and $\mathbf{Y}_\textrm{N}^{(i)} \in \{0, 1\}^T$, and $T$ is the total number of different types, our task aims to build a multi-label classification model to predict correct entity types for input contexts and mentions. For simplicity of notation, we define the complete entity type set as $\mathcal{T}=\{t_i\}_{i=1}^T$, where each $t_i$ is a type represented as a phrase, e.g., ``basketball player''. Therefore, each label vector $\mathbf{Y}_{\textrm{G}}^{(i)}$ has a corresponding type set $\mathcal{\tau}_{\textrm{G}}=\{t_j | Y_{\textrm{G}}^{(i,j)}=1, j=1, \cdots,T\}$, similarly for $\mathbf{Y}_{\textrm{N}}^{(i)}$ with $\mathcal{\tau}_{\textrm{N}}$. 

\subsection{Model Architecture}
Our distantly supervised approach consists of two major components: a denoising module and an entity typing module. The denoising module models label noise based on the perturbed gold labeled data and existing noisy labeled data from distant supervision. In particular, we characterize two kinds of entity typing noise: i) low coverage (low recall), and ii) wrong labeling (low precision). Using a unified noise modeling mechanism, we build a connection between ground truth labels, observed labels, and noise.  
With reliable noise modeling, we can recover high-quality labels for noisy data and further train a more accurate entity typing model. Below we provide details of each component.




\subsection{Noise Modeling}
Given a certain context and mention pair $(x,m)$, we assume the relation among gold label $\mathbf{y}_{\textrm{G}}$ and observed (noisy) label $\mathbf{y}_{\textrm{N}}$ is given by:
\begin{align}
\label{eq:noise-eq}
    \mathbf{y}_{\textrm{G}} = [\min(\mathbf{y}_{\textrm{N}} - \mathbf{e}, 1)]_+
\end{align}
where $[x]_+=\max(x, 0)$, $\mathbf{e} \in \{-1, 0, 1\}^T$ is the noise term, including causes to both false positive and false negative errors. For gold labeled data $\mathcal{D}_{\textrm{G}}$, $\mathbf{y}_{\textrm{G}}=\mathbf{Y}_{\textrm{G}}^{(i)}$ and $\mathbf{e}=0$. For noisily labeled data $\mathcal{D}_{\textrm{N}}$, $\mathbf{y}_{\textrm{G}}$ and $\mathbf{e}$ are unknown. Our denoising aims at recovering a more trustworthy label $\mathbf{y}_{\textrm{G}}$ from its noisy observation $\mathbf{y}_{\textrm{N}}$ by subtracting $\mathbf{e}$. 

Figure~\ref{fig:architecture} illustrates the workflow of our denoising model. The noise model $\mathcal{N}_{\theta}(x,m, \mathbf{\tau})$ is a neural network model parameterized by $\theta$, which takes the query sentence $x$ with the target entity mention $m$ as well as the current assigned (noisy) type set $\mathbf{\tau}$ as input, and outputs the noise measure $\mathbf{e}=\mathcal{N}_{\theta}(x,m, \mathbf{\tau})$. 
By Eq (\ref{eq:noise-eq}), it is relatively easy to conclude that $e_i \rightarrow 0$ indicates no change in the corresponding type assignment for type $t_i$. Similarly, $e_i \rightarrow 1$ indicates changing the type assignment towards negative, and when $e_i \rightarrow -1$ means changing the type assignment towards positive. 

We use BERT~\cite{devlin2019bert} model to build $\mathcal{N}_{\theta}(.)$. Specifically, BERT jointly encodes input context, target mention as well as current assigned entity type set to $d$ dimensional vector for each token and we extract the vector corresponding to the first token [CLS] as a pooled representation of the input as $\textrm{Embed}(x, m, \tau) = \textrm{BERT}_{\textrm{CLS}}(\textrm{Joint}(x, m, \tau))$. 

To joint the context $x$, mention $m$ and current assigned entity types $\tau$ in an entity-aware manner, we first utilize the special tokens preserved in BERT to indicate the positions of the target entity mention in $x$. Specifically, we insert [E0]/[/E0] at the beginning/ending of the target mention $m$. Following the BERT convention, we add special tokens [CLS] and [SEP] into the spans of context text and the entity type text spans. To encode the assigned type set $\tau$, we concatenate the type's plain text after query $x$. Since there is no sequence order between types, for type phrases, the position ids of all the tokens in type phrase spans are set to be the length of encoded $x$. Hence $\textrm{Joint}(x, m, \tau)$ is defined as:
\begin{align*}
    \textrm{Joint}(x, m, \tau) = & \textrm{[CLS]} w_1, ..., \textrm{[E0]} w_p, ..., w_q \textrm{[/E0]}, \\ 
    & ..., w_n \textrm{[SEP]} t_i, ..., t_j \textrm{[SEP]},
\end{align*}
where $w_p, \cdots, w_q$ represents the tokens of mention $m$, and $t_i, \cdots, t_j$ are concatenated type phrases. 

The estimated noise $\mathbf{e}$ is calculated by appending a linear layer with $\tanh$ activation on $\textrm{Embed}(.)$:
\begin{align}
\label{eq:bert-denoise}
    \mathbf{e} = \tanh{(\mathbf{W} * \textrm{Embed}(x, m, \tau)+ \mathbf{b})},
\end{align}
where $\mathbf{W} \in \mathcal{R}^{d \times T}$ and $\mathbf{b} \in \mathcal{R}^T$ are trainable parameters. 

\subsubsection{Training Data for Noise Modeling}
We utilize both available small-scale gold data and large-scale distant supervision data to train our noise model. Below we use ultra-fine dataset~\cite{choi2018ultra} as the example. Other datasets can be processed similarly. 

\vspace{0.1in}
\noindent \textbf{Utilize Gold Labeled Data $\mathcal{D}_{\textrm{G}}$.}
We perturb the labels of $\mathcal{D}_{\textrm{G}}$ to mimic the low-recall and low-precision issues under distant supervision. First we analyze the average number of types in $\mathcal{D}_{\textrm{G}}$ and $\mathcal{D}_{\textrm{N}}$, respectively. In ultra-fine dataset~~\cite{choi2018ultra}, there are $5.4$ and $1.5$ types per gold example and DS example, respectively. 

To mimic the low-recall issue, for each instance $(x_{\textrm{G}}, m_{\textrm{G}}, \tau_{\textrm{G}})$ from $\mathcal{D}_{\textrm{G}}$, we randomly drop each type with a fixed rate $0.7$ independent of other types to produce a corrupted type set $\tau_{\textrm{G}}^{R}$. We denote the corrupted gold data with randomly dropped types as $\mathcal{D}_{\textrm{G}}^{\textrm{R}}$. Meanwhile, to mimic the low-precision issue, for each instance $(x_{\textrm{G}}, m_{\textrm{G}}, \tau_{\textrm{G}})$, we also randomly replace its gold entity type set $\tau_{\textrm{G}}$ to a random set $\tau_{\textrm{G}}^{P}$, where $\tau_{\textrm{G}}^{P}$ is randomly sampled from $\mathcal{D}_{\textrm{N}}$. Note that $\tau_{\textrm{G}}$ and $\tau_{\textrm{G}}^{P}$ may or may not have overlapping entity types. The non-overlapping replacement leads to a totally corrupted DS instance.  The overlapping replacement represents the partially correct labeled instance. We denote the corrupted gold data with randomly replaced labels as $\mathcal{D}_{\textrm{G}}^{\textrm{P}}$. Given the complete entity set $\mathcal{T}$, $\tau_{\textrm{G}}$ and $\tau_{\textrm{P}} \in \{\tau_{\textrm{G}}^{P}, \tau_{\textrm{G}}^{R}\}$, it is straightforward to construct multi-hot vector representations $\mathbf{y}_\textrm{G}$ (i.e., $\mathbf{y}_\textrm{G}=\mathbf{Y}_\textrm{G}^{(i)}$) and $\mathbf{y}_\textrm{P} \in \{0, 1\}^T$. Finally, we collect the combined perturbation dataset $\mathcal{D}_{\textrm{P}}=\mathcal{D}_{\textrm{G}}^{\textrm{P}} \cup  \mathcal{D}_{\textrm{G}}^{\textrm{R}}$. 

\vspace{0.1in}
\noindent \textbf{Utilize Distant Supervision Data $\mathcal{D}_{\textrm{N}}$.}
Although the perturbed dataset $\mathcal{D}_{\textrm{P}}$ could be large, the gold labeled dataset $\mathcal{D}_{\textrm{G}}$ per se is still small, which means the number of different query sentences in $\mathcal{D}_{\textrm{P}}$ is limited. Hence training the noise model~$\mathcal{N}_\theta(.)$~only~on $\mathcal{D}_{\textrm{P}}$ may be insufficient for satisfactory performance. 

Although distant supervision datasets are noisy and the noise is unknown, they still can provide weak supervision. Hence we use the available large-scale DS dataset $\mathcal{D}_{\textrm{N}}$ to better train $\mathcal{N}_\theta(.)$. Our motivation grounds on the study in~\citet{choi2018ultra} showing that removing any source of distant supervision data from the whole training set results in a significant performance drop of the entity typing model. In other words, DS data contains a significant amount of correctly assigned entity types. Inspired by the analysis in~\citet{choi2018ultra}, we argue that the estimated noise $\mathbf{e}$ on DS data should be sparse. The sparsity enables us to design a suitable loss function to use $\mathcal{D}_{\textrm{N}}$ in training $\mathcal{N}_\theta(.)$. 

\vspace{0.05in}
\subsubsection{Objective Function for Noise Modeling}
\vspace{0.05in}

Training the noise model $\mathcal{N}_\theta(.)$ on $\mathcal{D}_{\textrm{P}}$ is a supervised learning procedure, and we apply the binary cross-entropy loss on each entity type. We consider below loss function for one corrupted input $((x, m, \mathbf{y}_{\textrm{P}}, \tau_{\textrm{P}}), \mathbf{y}_{\textrm{G}})$ from $\mathcal{D}_{\textrm{P}}$: 
\begin{align}
\label{eq:DP-obj}
  &  J_{\mathcal{D}_{\textrm{P}}} = -\sum_{t=1}^T [\mathbf{y}_{\textrm{G}}^{(t)} \cdot \log \hat{y_t} + (1- \mathbf{y}_{\textrm{G}}^{(t)}) \cdot \log (1- \hat{y_t})]  \nonumber \\
& \hat{y_t} = [(\textrm{min}(\mathbf{y}_{\textrm{P}}^{(t)}- \mathcal{N}_\theta^{(t)}(x, m, \tau_{\textrm{P}}), 1)]_{+}
\end{align}

To utilize $\mathcal{D}_{\textrm{N}}$ on training $\mathcal{N}_\theta(.)$, we use $L_1$ norm regularization on the difference between predicted labels and observed (noisy) labels. $L_1$ norm enforces sparseness, which leads to zero noise on a certain entity types. Such a procedure makes our prediction partially consistent with observed noisy labels, which is reasonable since $\mathcal{D}_{\textrm{N}}$ contains a significant amount of correct labels~\cite{choi2018ultra}. The loss function on one instance $((x, m, \tau_{\textrm{N}}), \mathbf{Y}_{\textrm{N}}^{(i)})$ from $\mathcal{D}_{\textrm{N}}$ is as follows:
\begin{align}
\label{eq:DN-obj}
    J_{\mathcal{D}_{\textrm{N}}} = ||\mathbf{\hat{y}}- \mathbf{Y}_{\textrm{N}}^{(i)}||_1, 
\end{align}
where $\mathbf{\hat{y}} = [(\textrm{min}(\mathbf{Y}_{\textrm{N}}^{(i)}-\mathcal{N}_\theta(x, m, \tau_{\textrm{N}}), 1)]_{+}$. The overall objective function becomes:
\begin{align}
\label{eql:denoising_obj}
    J_{\textrm{denoising}} = J_{\mathcal{D}_{\textrm{P}}}+ \alpha* J_{\mathcal{D}_{\textrm{N}}}
\end{align}
where $\alpha \geq 0$ is the regularization parameter, which is set as a small value so that distant supervision data can provide weak supervision but without overwhelming the training procedure.

\subsection{Entity Typing Model}
\begin{figure}[!htb]
\centering
\includegraphics[width=3in]{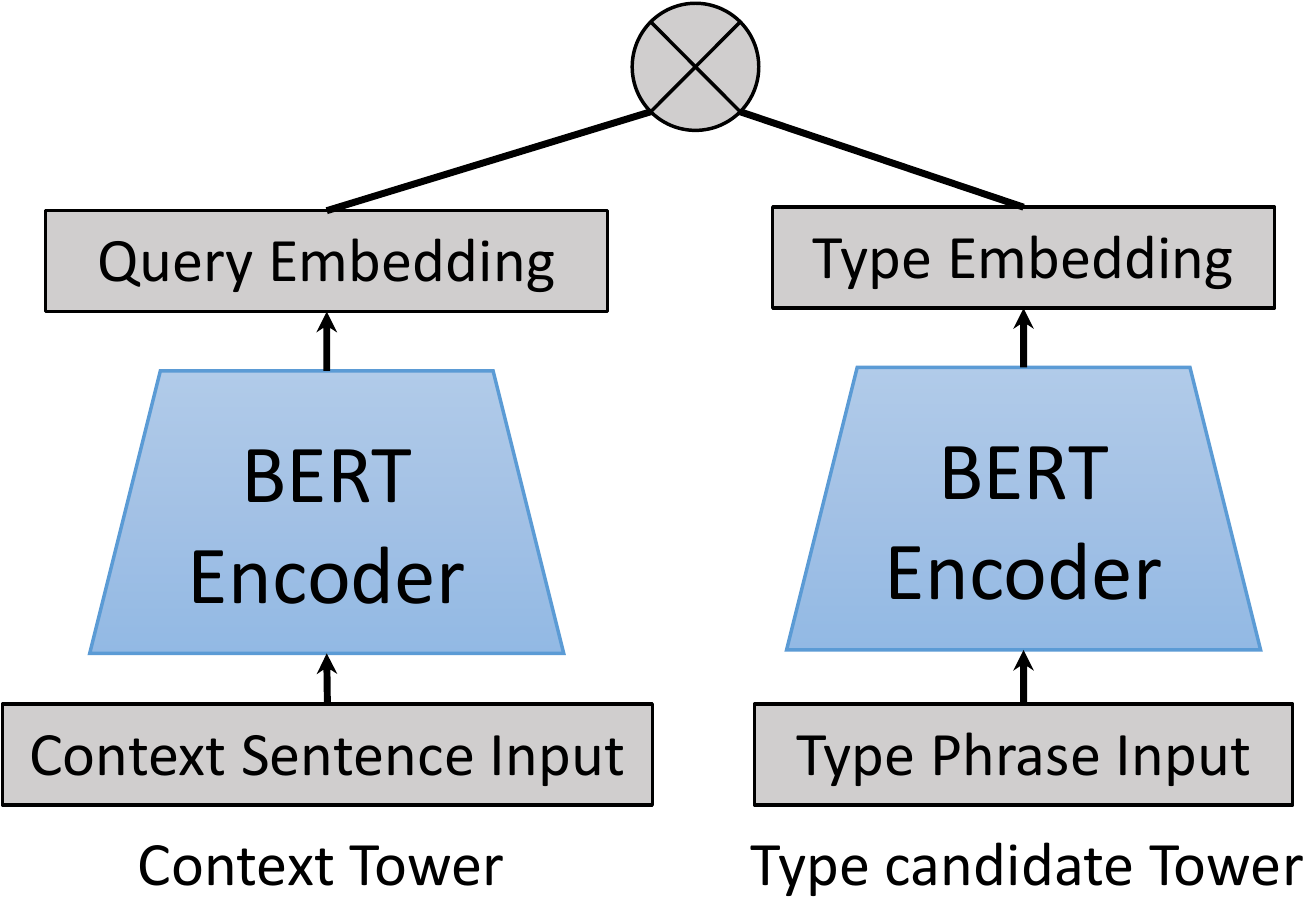} 
\vspace{-0.2in}
\caption{The architecture of entity typing model. } 
\label{fig:entityTyping}
\end{figure}

After training the noise model, we apply the learned model $\mathcal{N}_\theta(.)$ on DS data $\mathcal{D}_{\textrm{N}}$ to get the denoised dataset $\mathcal{D}_{\textrm{D}}$. We then use both $\mathcal{D}_{\textrm{G}}$ and $\mathcal{D}_{\textrm{D}}$ to train our entity typing model $\mathcal{M}_\phi(.)$ parameterized by $\phi$. Our entity typing model adopts the two-tower architecture, including the context tower and type candidate tower, as shown in Figure~\ref{fig:entityTyping}. 

In particular, the context tower takes the context sentence as input. We encode the sentence in an entity-aware manner using BERT model:
\begin{align*}
   & \textrm{Joint}_{\textrm{context}}(x) = \textrm{[CLS]} w_1, ..., \textrm{[E0]} w_p, ..., w_q \textrm{[/E0]}... \\
 &   \textrm{Embed}_{\textrm{context}}(x) = \textrm{BERT}_\textrm{CLS}(\textrm{Joint}_{\textrm{context}}(x))
\end{align*}

The candidate tower takes one entity type phrase as input. Again, we use another BERT model to encode the type phrase:
\begin{align*}
    \textrm{Joint}_{\textrm{candidate}}(t) &= \textrm{[CLS]}  w_1, ..., w_n \textrm{[SEP]} \\
    \textrm{Embed}_{\textrm{candidate}}(t) &= \textrm{BERT}_\textrm{CLS}(\textrm{Joint}_{\textrm{candidate}}(t))
\end{align*}
where $w_1, ..., w_n$ represents tokens of one type $t$. 

The final matching score $s(x, t)$ is computed as the inner product of the query embedding and the type embedding followed by a sigmoid activation:
\begin{align*}
    s(x, t) = \sigma(\textrm{Embed}_{\textrm{context}}(x)^T \textrm{Embed}_{\textrm{candidate}}(t))
\end{align*}
where $\sigma(.)$ is the sigmoid function, which maps the value into 0 to 1. In our entity typing model $\mathcal{M}_\phi$, we independently compute the matching score for each candidate type $t$. 

\vspace{0.1in}
\noindent \textbf{Objective function.}
Previous works~\citet{choi2018ultra, xiong2019imposing, onoe2019learning, onoe2021modeling, dai2021ultra} all adopt multi-task learning to handle the labeling noise, where they partition the labels into general, fine, and ultra-fine classes, and only treat an instance as an example for types of the class in question if it contains a label for that class. The multi-task objective avoids penalizing false negative types and can achieve higher recalls. In our work, since we already denoise and re-label the distant supervision data using our learned model $\mathcal{N}_\theta(.)$, we directly train the entity typing model using cross entropy loss without multi-task learning:
\begin{align}
\label{eq:typing-obj}
    J_{\textrm{typing}} &= -\sum_{t=1}^T [y_t \cdot \log \hat{y_t} + (1- y_t) \cdot \log (1- \hat{y_t})] \nonumber \\
    \hat{y_t} &= \mathcal{M}_\phi(x, t) 
\end{align}

\subsection{Iterative Training}
In our framework, the noise model $\mathcal{N}_\theta(.)$ and the entity typing model $\mathcal{M}_\phi$ are iterative trained as shown in Figure~\ref{fig:itertative}. We describe one training iteration for $\mathcal{N}_\theta(.)$ and $\mathcal{M}_\phi$ in the following: 

\begin{figure}[!htb]
\centering
\includegraphics[width=3in]{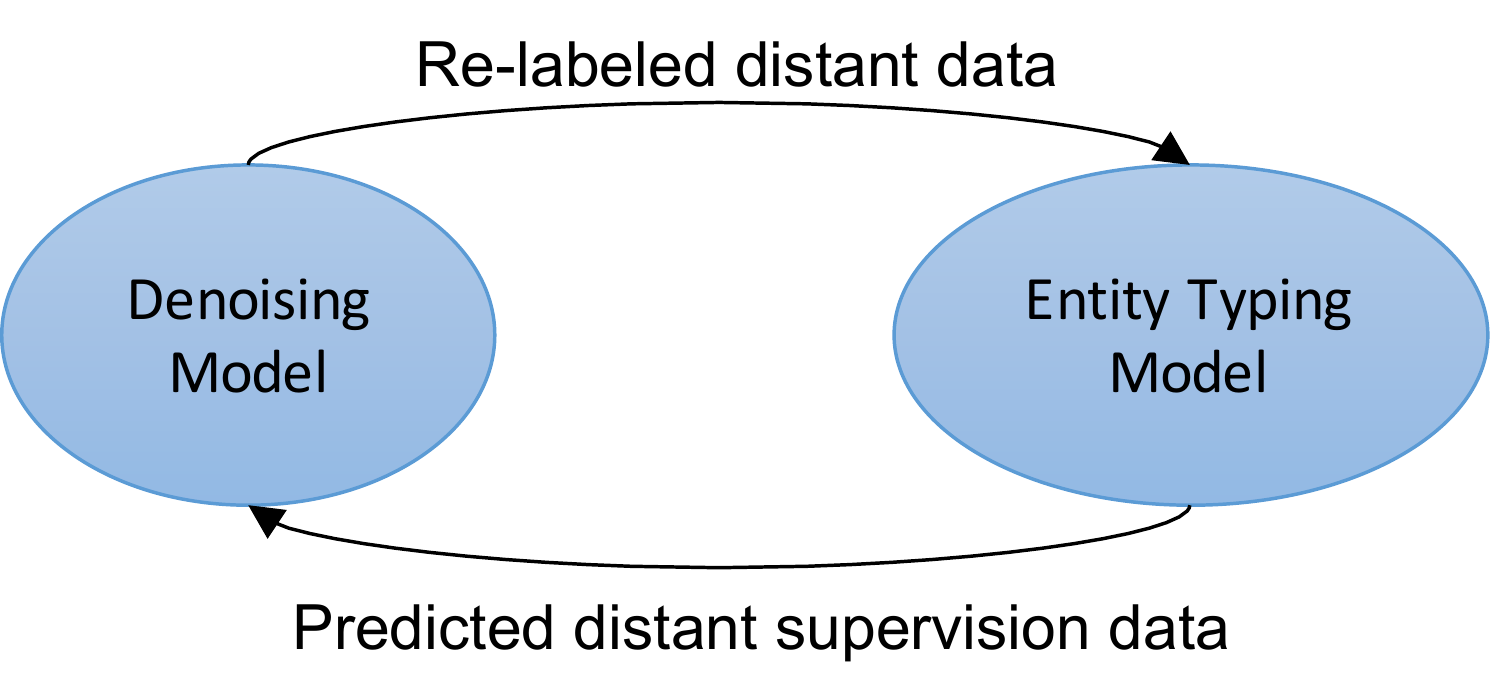} 
\vspace{-0.1in}
\caption{Illustration of the iterative training. } 
\label{fig:itertative}
\end{figure}

\paragraph{Updating $\mathcal{N}_\theta(.)$:} We train the noise model $\mathcal{N}_\theta(.)$ by Eq (\ref{eql:denoising_obj}) using the perturbed gold labeled dataset $\mathcal{D}_{\textrm{P}}$ and noisy dataset $\mathcal{D}^\prime$. At the first iteration, $\mathcal{D^\prime}$ is from the original distant supervision, $\mathcal{D^\prime}=\mathcal{D}_{\textrm{N}}$. After the first iteration, labels in $\mathcal{D}^\prime$ are re-calculated by applying current entity typing model $\mathcal{M}_\phi(.)$. Also, after each iteration, we increase the value of the weight $\alpha$ in Eq (\ref{eql:denoising_obj}). After noise modeling, we get the denoised dataset $\mathcal{D}_{\textrm{D}}$ by removing the noise calculated from applying $\mathcal{N}_\theta(.)$ on $\mathcal{D^\prime}$. 

\paragraph{Updating $\mathcal{M}_\phi(.)$: } We train the entity typing model $\mathcal{M}_\phi(.)$ by Eq (\ref{eq:typing-obj}) using gold dataset $\mathcal{D}_{\textrm{G}}$ and the latest denoised dataset $\mathcal{D}_{\textrm{D}}$. After training current $\mathcal{M}_\phi(.)$, we re-calculate the labels of distant supervision data, and get the updated DS dataset $\mathcal{D^\prime}$. We pass $\mathcal{D^\prime}$ to the next noise modeling iteration. 

\begin{table*}[t!]
  \centering
  \begin{tabular}{c|cccc|cccc} 
  \toprule
  \multirow{2}{*}{Model} & \multicolumn{4}{c|}{Dev} & \multicolumn{4}{c}{Test} \\
    \cline{2-9}
    {}& MRR & P & R & F1 & MRR & P & R & F1\\
    \midrule
    AttentiveNER & 22.1 & 53.7 & 15.0 & 23.5 & 22.3 & 54.2 & 15.2 & 23.7 \\
    Multi-task & 22.9 & 48.1 & 23.2 & 31.3 & 23.4 & 47.1 & 24.2 & 32.0 \\
    \midrule
    LabelGCN & 25.0 & \textbf{55.6} & 25.4 & 35.0 & 25.3 & \textbf{54.8} & 25.9 & 35.1 \\
    BERT & {-} & 51.6 & 32.8 & 40.1 & {-} & 51.6 & 33.0 & 40.2 \\
    Filter+Relabel & {-} & 50.7 & 33.1 & 40.1 & {-} & 51.5 & 33.0 & 40.2 \\
    VectorEmb & {-} & 53.3 & 36.7 & 43.5 & {-} & 53.0 & 36.3 & 43.1 \\
    BoxEmb & {-} & 52.9 & 39.1 & 45.0 & {-} & 52.8 & 38.8 & 44.8 \\
    MLMET$^{*}$ & 29.0 & 53.6 & 39.4 & 45.4 & 29.2 & 53.4 & 40.5 & 46.1 \\
    \midrule
    Ours & \textbf{30.3} & 52.8 & 41.7 & 46.6 & \textbf{30.9} & 53.4 & 41.9 & 47.0 \\ 
    Ours+ \textit{thresholding} & \textbf{30.3} & 50.8 & \textbf{43.7} & \textbf{47.0} & \textbf{30.9} & 51.2 & \textbf{43.7} & \textbf{47.3} \\ 
  \bottomrule
  \end{tabular}
      \caption{Comparison with baseline models on the UFET dataset. All the baseline results are from their papers. ``-'' means no report. Best results with statistical significance are marked in bold (one-sample t-test with $p < 0.05$). ``*'' means we reproduced the results based on the public released code and dataset. }
  \label{table:ultrafine_result}
\end{table*}

\vspace{0.1in}
\section{Experiments}

\subsection{Experimental Setup}

\paragraph{Datasets}

Our experiments mainly focus on the Ultra-Fine entity typing (UFET) dataset, which has 10,331 labels. The distant supervision training set is annotated with heterogeneous supervisions based on KB, Wikipedia, and headwords, resulting in about 25.2M training samples. This dataset also includes around 6,000 crowdsourced samples equally split into training, validation, and test set. 

In addition, we investigate on OntoNotes dataset, which is a widely used benchmark for fine-grained entity typing systems. The initial training, development, and test splits contain 250K, 2K, and 9K examples, respectively.~\citet{choi2018ultra} augmented the training set to include 3.4M distant supervision examples. To train our noise model, we further augment the training data using the 2,000 training crowdsourced samples from the UFET dataset. We map the labels from ultra-fine types to OntoNotes types. Most OntoNote’s types can directly correspond to UFET's types (e.g., ``doctor'' to ``/person/doctor''). We then expand these labels according to the ontology to include their hypernyms (e.g., ``/person/doctor'' will also generate ``person''). 

\vspace{0.1in}
\noindent\textbf{Baselines.}
For the UFET dataset, we compare with 

\vspace{0.1in}

\noindent1) AttentiveNER~\cite{shimaoka2016attentive}; 

\noindent2) Multi-task model~\cite{choi2018ultra}, which is proposed together with the UFET data; 

\noindent3) LabelGCN~\cite{xiong2019imposing}; 

\noindent4) BERT~\cite{onoe2019learning}, which was first introduced as a baseline; 

\noindent5) Filter+Relabel~\cite{onoe2019learning}; 

\noindent6) Vector Embedding~\cite{onoe2021modeling};

\noindent7) Box Embedding~\cite{onoe2021modeling}; 

\noindent8) MLMET \cite{dai2021ultra}.

\vspace{0.1in}

For experiments on OntoNotes, additionally, we compare with AFET~\cite{ren2016afet}, LNR~\cite{ren2016label}, and NFETC~\cite{xu2018neural}.

\vspace{0.1in}\noindent\textbf{Evaluation Metrics.}\ For the UFET dataset, we report the mean reciprocal rank (MRR), macro precision(P), recall (R), and $F_1$. As P, R and $F_1$ all depend on a chosen threshold on probabilities, we tune the threshold on the validation set from 50 equal-interval thresholds between 0 and 1 and choose the optimal threshold which can lead to the best F1 score. Then, we use the found optimal threshold for the test set. Also, we plot the precision-recall curves, which are the more transparent comparison. For the OntoNotes dataset, we report the standard metrics used by baseline models: accuracy, macro, and micro F1 scores. 

\vspace{0.1in}\noindent\textbf{Implementation Details.}\ To train models on UFET dataset, all the baselines adopt the multi-task loss proposed in~\citet{choi2018ultra}. For our model, we use the standard binary cross-entropy (BCE) losses in Eq (\ref{eql:denoising_obj},~\ref{eq:typing-obj}). 
We carefully tune $\alpha$ from $[0.05, 0.1, 0.25, 0.5, 0.75, 1]$ and set it to $0.25$ based on validation set. We use ``BERT-base-uncased'' to initialize Bert encoder weights, and set dropout rate to 0.1. We use Adam optimizer~\cite{kingma2015adam} with learning rate $3e-5$. We repeat our experiments five times and report the average metrics on the test set.  

MLMET results are reproduced using the public released code and data, not directly taken from their paper. For all our ultra-fine entity typing experiments, we consider 25.2M distant supervision training samples and 6,000 crowdsourced samples equally split into training, validation, and test set. While, the original MLMET also consider additional 3.7M pronoun mentions dataset from EN Gigaword.

\subsection{Evaluation Results}

\noindent \textbf{Evaluation on UFET Dataset.}
We report the comparison results on UFET in Table~\ref{table:ultrafine_result}. MRR score is independent with threshold choices. For F1 score, we apply threshold-tuning, which further improves the F1 score on both the development and test sets. In terms of MRR and F1, our model outperforms baseline methods by a large margin, especially on the test set. We can see that recall is usually lagging behind precision by a large margin for most baseline models. It is because that these baselines easily correctly predict the nine general types but have difficulty predicting the large number of fine-grained and ultra-fine types correctly. On the other hand, our model can balance the precision and recall scores well even without threshold-tuning. The ``thresholding'' sacrifices the precision and tunes towards recall to lead to a higher F1 score. 

\begin{figure}[!htb]

\includegraphics[width=3.2in]{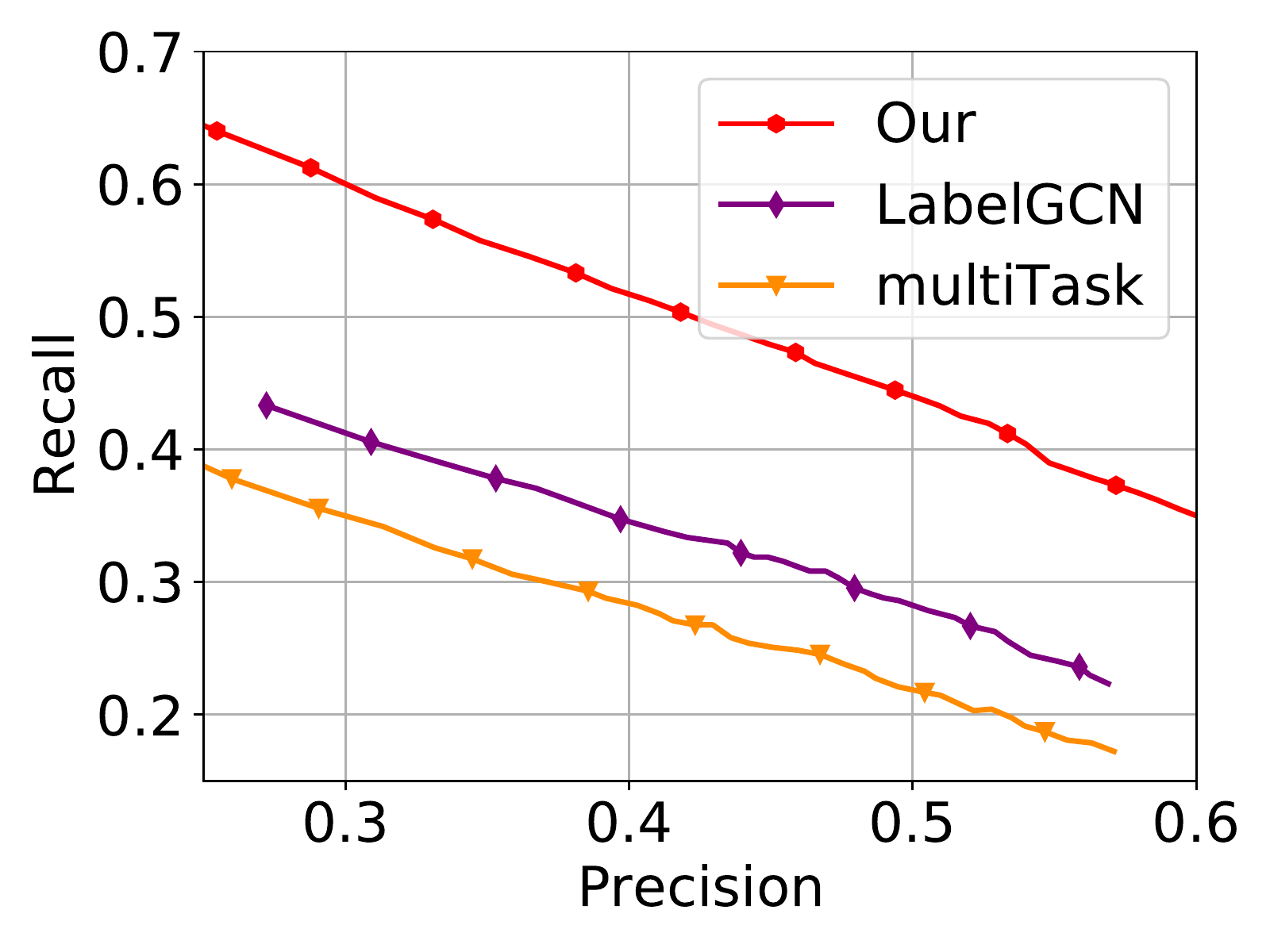} 
\vspace{-0.2in}
\caption{Precision-recall curves on UFET dev set.} 
\label{fig:PR_curve}
\end{figure}

For a more transparent comparison, we show the precision-recall curves in Figure~\ref{fig:PR_curve}. These data points are based on the performance on the development set given by 50 equal-interval thresholds between 0 and 1. We can see there is a clear margin between our model v.s. LabelGCN and multiTask models. With higher recalls or more retrieved types, achieving high precision requires being accurate on fine-grained and ultra-fine types, which are often harder to predict.

\vspace{0.1in}\noindent \textbf{Evaluation on OntoNotes Dataset.}
We report the comparison results on OntoNotes in Table~\ref{table:onto_result}. Baseline models including AttentiveNER, AFET, LNR, and NFETC explicitly use the hierarchical type structures provided by the OntoNotes ontology. While other baselines and ours do not consider the type hierarchy and treat each type as a free-form phrase. From Table~\ref{table:onto_result}, we can see that our model significantly outperforms other baselines on all the metrics, especially on the accuracy metric. 

\begin{table}[t]
    \centering
    \begin{tabular}{c|ccc}
    \toprule
    Model & Acc & Mac-F1 & Mic-F1 \\
    \midrule
    AttentiveNER & 51.7 &  71.0 &  64.9 \\
    AFET & 55.1 & 71.1 & 64.7 \\
    LNR & 57.2 & 71.5 & 66.1 \\
    NFETC & 60.2 & 76.4 & 70.2 \\
    \midrule
    Multi-task & 59.5 & 76.8 & 71.8 \\
    LabelGCN & 59.6 & 77.8 & 72.2 \\
    BERT & 51.8 & 76.6 & 69.1 \\
    Filter+Relabel & 64.9 & 84.5 & 79.2 \\
    MLMET & 67.4 & \textbf{85.4} & 80.4 \\
    \midrule
    Ours & \textbf{70.0} & \textbf{85.4} & \textbf{80.9} \\
    \bottomrule
    \end{tabular}
    \caption{Comparison results on OntoNotes. Best results with statistical significance are marked in bold.}
    \label{table:onto_result}
\end{table}


\subsection{Analysis and Ablation Study}

\subsubsection{Utility of Iterative Training}

First, we analyze the effectiveness of iterative training. We report the test results of our model and Filter+Relabel model under different iteration steps. In the original Filter+Relabel model, the filter and relabel functions are trained on the gold data only. To make the Filter+Relabel model take advantage of the iterative training, we relabel all the DS data by the trained entity typing model after each iteration. Then we leverage the current filtering function to evaluate the relabeled DS samples and filter out the high-quality DS samples. Then we joint the high-quality DS samples with gold data to train filter and relabel functions in the next iteration.

\begin{table}[ht]
\vspace{0.1in}
  \centering
  \begin{tabular}{c|cccc} 
  \toprule
    Iteration & Model  & P & R & F1 \\
    \midrule
    \multirow{2}{*}{1} & Ours & 46.6 & 46.4 & 46.5 \\
    {} & Filter+Relabel & 50.7 & 33.1 & 40.1 \\
        \midrule
    \multirow{2}{*}{2} & Ours & 48.0 & 46.1 & 47.0 \\
    {} & Filter+Relabel & 52.3 & 35.0 & 41.9 \\
        \midrule
     \multirow{2}{*}{3} & Ours & 51.2 & 43.7 & 47.3 \\
     {} & Filter+Relabel & {52.9} & {35.1} & 42.2 \\
  \bottomrule
  \end{tabular}
    \vspace{-1mm}
      \caption{Test results of our model and Filter+Relabel on UFET under different iteration training steps.}
  \label{table:analysis_iteration}
\end{table}

Table~\ref{table:analysis_iteration} shows that with the step of iteration increasing, the overall performance of both models get better. This proves the significance of iterative training. With the iteration step increasing, the DS data becomes less noisy. Also, we can see at every iteration step, our model is much better than Filter+Relabel, which proves that modeling the noise instead of the label is a better choice. 

\vspace{0.05in}
\subsubsection{Effectiveness of Noise Modeling}
\vspace{0.05in}

Since noise modeling's output directly impacts the final entity typing performance, we also quantify the performance of trained $\mathcal{N}_\theta(.)$ on held-out gold-labeled dev set and perturbed gold dev set to mimic the low-recall and low-precision scenarios. 

\begin{table}[ht!]
    \centering
    \begin{tabular}{c|cc}
        \toprule
        Data & Filter+Relabel  & Ours  \\
        \midrule
        Gold Dev & 87.9 & \textbf{94.2} \\
        Low-recall set & {54.5} & \textbf{55.8} \\
        Low-precision set & {33.0} & \textbf{35.5} \\
        \bottomrule
    \end{tabular}
    \caption{F1 scores of noise modeling on three datasets. }
    \label{table:quantify_denoise}
\end{table}

We report the $F_1$ score on these three datasets in Table~\ref{table:quantify_denoise}. To fairly compare with Filter+Relabel, we re-implement it using BERT as the backbone. Our denoising module generates more accurate prediction and is more robust to noise. 

\vspace{0.05in}
\subsubsection{Consistency Analysis}
\vspace{0.05in}

We investigate whether our model can predict type relations in a consistent manner. Following the evaluation in~\citet{onoe2021modeling}, we conduct the analysis on the UFET dev set. We count the number of occurrences for all subtypes in 30 (supertype, subtype) pairs listed in~\citet{onoe2021modeling}. Then, for each subtype, we count how many times its corresponding supertype is also predicted. Finally, the accuracy (acc) is the ratio of predicting the corresponding supertype when the subtype is exhibited.

\begin{table}[ht]
    \centering
    \begin{tabular}{c|ccc}
    \toprule
    Model & \# (sup, sub) & \# sub & acc \\
    \midrule
    VectorEmb & {1451} & 1631 & 89.0 \\
    BoxEmb & {1851} & 1997 & 92.7 \\
    Ours & \textbf{2514} & \textbf{2674} &  \textbf{94.0} \\
    \bottomrule
    \end{tabular}
    \caption{Consistency: accuracy evaluated on the 30 (supertype, subtypes) pairs. }
    \label{table:consistency}
\end{table}

Table~\ref{table:consistency} reports the count and accuracy of the 30 (supertype, subtype) pairs, where the \#(sup, sub) column shows the number of pairs found in the predictions, \# sub column shows the number of subtypes found in the predictions. Our model achieves a higher count and accuracy. Intuitively, a higher count indicates a higher recall of the model. A higher accuracy proves that although the supertype-subtype relations are not strictly defined in the training data, our model still captures the correlations. 

\vspace{0.05in}
\subsubsection{Ablation Study}
\vspace{0.05in}

To prove the effectiveness of our denoising mechanism for the entity typing task, we conduct an ablation study on the UFET dev set and show the results in Table~\ref{table:noising_ultrafine}. We study three model variants, including i) full model w/o denoising, where we train entity typing model on gold data and DS data; ii) full model w/o denoising or DS data, where we train entity typing model only on gold data; and iii) full model w/o cross-attention between the input context and assigned entity type phrases when modeling noise.

\begin{table}[ht]
{\hspace{-0.2in}
  \begin{tabular}{c|cccc} 
  \toprule
    Model & MRR & P & R & F1 \\
    \midrule
    Full model & \textbf{30.3} & \textbf{50.7} & \textbf{43.5} & \textbf{46.8} \\
    w/o denoising & 27.2 & 43.7 & 39.2 & 41.3 \\
    w/o denoising or $\mathcal{D}_{\textrm{N}}$ & 28.3 & 45.3 & 39.7 & 42.3 \\
    w/o cross-attn. & 29.9 & 47.1 & 43.9 & 45.4 \\
  \bottomrule
  \end{tabular}
}
      \caption{Ablation study on the UFET test set.}
  \label{table:noising_ultrafine}
\end{table}



From Table~\ref{table:noising_ultrafine}, we first observe that directly training an entity typing model without our denoising mechanism results in a significant performance drop. Second, we see that introducing more distant supervision data $\mathcal{D}_{\textrm{N}}$ can improve the overall entity typing performance. Finally, joining the context and assigned types and introducing self-attention~\cite{vaswani2017attention} further improves $F_1$ score by $~1.6\%$. Therefore, when designing the denoising model, it is necessary to fully explore the dependency between the context sentence and the assigned type set instead of a simple sum-pooling in Filter+Relabel~\cite{onoe2019learning}.

\subsection{Case Study}
To better explore the effectiveness of our denoising model, we show two case studies from the DS development set. We show all types with at least one score over the threshold of 0.5, or is annotated true by distant supervision in Table~\ref{tab:case1} and Table~\ref{tab:case2}. The target entity mentions are underlined within brackets. 

\vspace{0.1in}
\noindent \textbf{Case Study S1: } For the context ``My grandfather joined an [\underline{artillery regiment}] with the Canadian Expeditionary Force and then set off to fight...'', as shown in Table~\ref{tab:case1}, DS label misses the type ``group'', our denoising mechanism successfully identify the missing type, but Filter+Relabel~\cite{onoe2019learning} fails. 
\begin{table}[ht!]
    \centering
    \begin{tabular}{c|ccc}
    \toprule
    Type & DS & Filter+Relabel  & Ours  \\
    \midrule
    \texttt{group} & 0 & 0.0 & 0.56 \\
    \texttt{artillery} & 1 & 0.67 & 0.83 \\
    \texttt{regiment} & 1 & 0.19 & 0.73\\
    \bottomrule
    \end{tabular}
    \vspace{-2mm}
    \caption{Case study on DS instance S1, which is labeled by head words. }
    \label{tab:case1}
    \vspace{0.2in}
    \centering
    \begin{tabular}{c|ccc}
    \toprule
    Type & DS & Filter+Relabel & Ours \\
    \midrule
    \texttt{person} & 1 & {0.00} & {0.05} \\
    \texttt{engineer} & 1 & {0.54} & {0.46} \\
    \texttt{writing} & 1 & {0.92} & {0.88} \\
    \bottomrule
    \end{tabular}
    \vspace{-2mm}
    \caption{Case study on DS instance S2, which is labeled by entity linking. }
    \label{tab:case2}
\end{table}



\vspace{0.1in}
\noindent \textbf{Case Study S2: } For the context ``Richard Schickel, writing in [\underline{Time magazine}] gave a mixed review, ...'', DS wrongly assigned ``person'' and ``engineer'' to the entity ``Time magazine''. Both Filter+Relabel and our work successfully lower the probability scores of the wrong types.

\section{Conclusion}

In this paper, we aim to improve the performance of ultra-fine entity typing by explicitly denoising the DS data. Noise modeling is our key component to denoise, where the model fully explores the correlation between the query context and assigned noisy type set, and outputs the estimated noise. To train the noise model, we perturb on the small-scale gold dataset to mimic the noise distribution on DS instances. Furthermore, we utilize the large-scale DS data as weak supervision to train our noise model. The entity typing model is then trained on the gold data set and denoised DS dataset. Experimental results empirically prove the effectiveness of our method on handling distantly supervised ultra-fine entity typing. 

\bibliographystyle{acl_natbib}
\bibliography{refs_scholar}


\end{document}